# Sensitivities: An Alternative to Conditional Probabilities for Bayesian Belief Networks

**Alexander V. Kozlov**
Department of Applied Physics
Stanford University
Stanford, CA 94305
email: *alexvk@cs.stanford.edu*

**Jaswinder Pal Singh**
Department of Computer Science
Princeton University
Princeton, NJ 08544
email: *jps@cs.princeton.edu*

## Abstract

We show an alternative way of representing a Bayesian belief network by sensitivities and probability distributions. This representation is equivalent to the traditional representation by conditional probabilities, but makes dependencies between nodes apparent and intuitively easy to understand. We also propose a QR matrix representation for the sensitivities and/or conditional probabilities which is more efficient, in both memory requirements and computational speed, than the traditional representation for computer-based implementations of probabilistic inference. We use sensitivities to show that for a certain class of binary networks, the computation time for approximate probabilistic inference with any positive upper bound on the error of the result is independent of the size of the network. Finally, as an alternative to traditional algorithms that use conditional probabilities, we describe an exact algorithm for probabilistic inference that uses the QR-representation for sensitivities and updates probability distributions of nodes in a network according to messages from the neighbors.

## 1 INTRODUCTION

A belief network is a directed acyclic graph (DAG) together with a set of conditional probabilities associated with each node.[1] Given a probability distribution for the parents of a node, we can calculate the probability of the node by applying the chain rule of conditional probabilities.

Belief networks are now a well-established representation of knowledge and are used for forecasting, diagnosis, planning, learning, vision, and natural language processing. Several practical networks have been constructed, the most cited being MYCIN, a knowledge base for infectious disease diagnosis [Shortliffe, 1976], and PROSPECTOR, a system to aid geologists in mineral exploration [Campbell, 1982]. One of the largest networks built so far is QMR-DT, a medical knowledge database consisting of more than 4000 evidence nodes (test results, or facts about a patient) and 600 disease nodes [Shwe et al., 1991; Middleton et al., 1991].

A belief network contains two types of information: information about the conditional independence of different sets of nodes, and information about dependencies between nodes. The first is mostly encoded in the graphical structure of the belief network [Shachter, 1986; Howard, 1990; Shachter, 1991], and the second, in the numerical values of conditional probabilities [Pearl, 1988; Neapolitan, 1990]. Both types of information are important for applications. To make this information useful we have to have a method to extract it. The process of extracting information from a belief network is called probabilistic inference. A typical probabilistic inference is a process to answer a query about the probability of some nodes, called *query* nodes, given evidence about other nodes, called *instantiated* nodes. In general, probabilistic inference is *NP*-hard [Cooper, 1990].

Different algorithmic approaches have been proposed for probabilistic inference. The most successful technique for singly connected networks is called local message passing, since the inference algorithm can be conveniently represented as a process of passing information up and down the network by messages between adjacent nodes. In a general network with loops, direct application of the local message passing technique does not work, and more complex algorithms have been proposed. Two of the most efficient exact inference algorithms are the Lauritzen-Spiegelhalter (LS) algorithm [Lauritzen and Spiegelhalter, 1988] and an optimal factoring algorithm based on the symbolic probabilistic inference (SPI) approach [Li and D'Ambrosio, 1994]. Both of the algorithms have a time complexity that is exponential to the size of a general network.

Both of the above-mentioned algorithms have to use a lot of memory due to the internal representation of clique potentials (LS) or partial sums (SPI) as well as conditional probabilities. Since the memory requirements and computation time increase exponentially, it is very hard to go beyond some limit. Some combinations of the clique potentials, sums, or conditional probabilities, however, might be more important than the others for the answer to a query. This paper is a step towards finding these combinations and towards a future algorithm for an approximate probabilistic inference.

---

[1] Nodes without parents are not conditioned on any event; we consider conventional probabilities as conditional probabilities conditioned on an empty set of nodes.

In Section 2 we describe the notations used in the paper. In Section 3 we introduce the concept of *sensitivity*. A sensitivity can be intuitively understood as a linear transformation from the change in the probability of one node to the change in the probability of another node. We prove a number of properties and provide a *QR-representation* for sensitivities. This representation is efficient when the rank of sensitivity and/or probability matrix is low, as it is likely to be for the compound nodes that result from the conversion of an arbitrary network to a tree network. We also show that for binary nodes, an important parameter that defines the computational complexity of an approximate probabilistic inference might be the *difference* between conditional probabilities for different parent instantiations, and prove that for certain class of binary networks computation time for an approximate probabilistic inference with the *exact* upper bound on the relative error of the result *does not depend* on the size of the network. In Section 4 we provide two *algorithms for probabilistic inference* in trees of multiply-valued or compound nodes. The algorithms update prior probability distributions of the nodes according to the messages from the neighbors. Unlike the LS algorithm, we do not have to propagate messages throughout the whole network, but *only* along the chains of nodes between the instantiated nodes and the query node. Finally, in Section 5, we present some examples to illustrate the use of sensitivities in performing probabilistic inference.

## 2 NOTATIONS

If a node in a belief network can take two and only two values, *false* or *true*, we call it a *binary node*. We denote a binary node by a small letter $x$, with a possible subscript when further distinction is required. For binary nodes we use the short form $p(i)$ to mean the probability of node $x_i$ being in the state *true* and $p(\bar{i})$ to mean the probability of node $x_i$ being in the state *false*, i.e. $p(i) \equiv p(x_i = true)$ and $p(\bar{i}) \equiv p(x_i = false)$. Thus, $p(i|\bar{j})$ denotes the conditional probability of node $x_i$ having the value *true* when node $x_j$ has the value *false*.

We denote a *multiply-valued node* or a *set of binary nodes* by a capital letter $X$, again with a possible subscript. A superscript, if it appears, denotes a particular state of the multiply-valued node. The same notation for a multiply-valued node and a set of binary nodes is a reflection of the fact that they are equivalent. The set of nodes has a mutually exclusive and exhaustive set of values that can be represented as a multiply-valued node. We denote the number of nodes in a set as $N(X_i)$ and the total number of states as $|X_i|$ (for a set of binary nodes represented as a multiply-valued node $|X_i| = 2^{N(X_i)}$).

The *conditional probability* of a node is a set of numbers, each number being the probability that the node takes a particular one of its possible values given that the parent nodes take a particular one of their values. If a node $X_i$ with a set of possible values of size $|X_i|$ has only one parent $X_j$ with a set of possible values of size $|X_j|$, then a conditional probability $p(X_i|X_j)$ is a set of $|X_i| \times |X_j|$ numbers. A conditional probability is a normalized set of numbers, that is:

$$\sum_p p(X_i^p|X_j^q) = 1, \qquad \text{for all } X_j^q. \qquad (1)$$

We associate a conditional probability with every node in a belief network.

The *joint probability* of a particular combination of values of all nodes in a network is the product of all conditional probabilities in the network when the nodes assume these values:

$$p(X_1, X_2, \ldots, X_n) = \prod_{i=1}^n p(X_i|\mathbf{Pa}(X_i)), \qquad (2)$$

where $\mathbf{Pa}(X_i)$ denotes the set of parents of node $X_i$ in the network. We call the set of all of all joint probabilities of a network the *joint probability distribution* of the network. A belief network can thus be viewed as an efficient representation of the joint probability distribution by a product of conditional probabilities. To calculate the probability of a particular node one has to sum the joint probability distribution over all possible values of all the other nodes. We call the set of all $|X|$ probabilities of a node $X$ the *probability distribution* of the node $X$. Different algorithms for probabilistic inference can be seen as different strategies to sum up the joint probability distribution to get the probability distribution of a node.

To perform probabilistic inference with a set of instantiated nodes $I$ we could instantiate one node at a time, each time obtaining a new network with new conditional probabilities and, possibly, new edges. We call such a method an *incremental instantiation* with the set $I$. We distinguish between a *prior* probability distribution, the probabilities before any one of the nodes in a belief network was instantiated, and a *posterior* probability distribution, the probabilities after some nodes have been instantiated. We denote the prior probability distribution of a node $X_i$ with superscript (0): $p^{(0)}(X_i)$. The posterior probability distribution can refer to different number of incremental instantiations. We denote a posterior probability distribution of a node $X_i$ after $k$ consecutive node instantiations as $p^{(k)}(X_i)$. Other parameters of a network can have instantiation superscripts too.

In general, probabilistic inference can be reduced to two basic operations: *node reduction* and *arc reversal*. Every time we sum joint probability over all values of a node, we will say that that node has been reduced. For example, the joint probability distribution for a chain of three binary nodes $x_1 \rightarrow x_2 \rightarrow x_3$ is $p(x_1, x_2, x_3) = p(x_3|x_2)p(x_2|x_1)p(x_1)$. The summation over node $x_2$ gives:

$$p(x_1, x_3) \equiv \sum_{x_2} p(x_1, x_2, x_3)$$
$$= \Big(p(x_3|2)p(2|x_1) + p(x_3|\overline{2})p(\overline{2}|x_1)\Big)p(x_1)$$
$$= p(x_3|x_1)p(x_1). \quad (3)$$

The reduction of node $x_2$ from a binary node chain $x_1 \rightarrow x_2 \rightarrow x_3$ takes 8 multiplications and 4 summations considering all combinations of $x_1$ and $x_3$. Arc reversal [Shachter, 1986] is the operation of changing the representation of a joint probability distribution. In the example above, $p(x_1, x_3)$ is represented as $p(x_3|x_1)p(x_1)$. We can also represent $p(x_1, x_3)$ as $p(x_1|x_3)p(x_3)$ by the application of Bayes' rule:

$$p(x_1|x_3) \equiv \frac{p(x_1, x_3)}{\sum_{x_1} p(x_1, x_3)} = \frac{p(x_3|x_1)p(x_1)}{\sum_{x_1} p(x_3|x_1)p(x_1)}. \quad (4)$$

The arc reversal for a binary node chain $x_1 \rightarrow x_3$ takes 4 multiplications, 2 summations, and 4 divisions, again, considering all combinations of $x_1$ and $x_3$. We will show that the introduction of sensitivities simplifies both node reduction and arc reversal, and hence probabilistic inference.

In this paper, we often represent conditional probabilities and probability distributions as matrices for economical reasons. We denote a probability distribution $p(X_i)$ of a node $X_i$ as a $|X_i| \times 1$ column $\mathbf{P}_i$ and a conditional probability $p(X_i|X_j)$ of a node $X_i$ with respect to the node $X_j$ as a $|X_i| \times |X_j|$ matrix $\mathbf{P}_{ij}$. We use parentheses when indexing into a matrix: the element in the $p$-th row and $q$-th column of the matrix $\mathbf{A}_{ij}$ is denoted by $(\mathbf{A}_{ij})_{pq}$. We use standard notations of $\mathbf{I}$ for the identity matrix and $\mathbf{E}$ for the matrix consisting of ones. The diagonal matrix with the values of probability distribution $\mathbf{P}_i$ on the main diagonal is denoted by $\mathbf{\Lambda}_i$.

## 3 SENSITIVITIES

### 3.1 DEFINITION AND QR-REPRESENTATION

The *sensitivity* of a multiply-valued node $X_i$ with respect to a multiply-valued node $X_j$ is a matrix $\mathbf{S}_{ij}$ of size $|X_i| \times |X_j|$:

$$(\mathbf{S}_{ij})_{pq} \equiv p(X_i^p|X_j^q) - \frac{\sum_s p(X_i^p|X_j^s)}{|X_j|} \quad (5)$$
$$= \left(\mathbf{P}_{ij}(\mathbf{I} - \frac{1}{|X_j|}\mathbf{E})\right)_{pq}.$$

This transformation from the conditional probability $p(X_i|X_j)$ to sensitivity $\mathbf{S}_{ij}$ removes the subspace of a vector $(1, 1, \ldots, 1)$ from the row space of the conditional probability matrix $\mathbf{P}_{ij}$. Since at least some of the conditional probabilities are positive, the rank of the $\mathbf{S}_{ij}$ matrix is always one less than the rank of the conditional probability matrix $\mathbf{P}_{ij}$ (if all probabilities $p(X_i^p|X_j^q)$ for a given state $X_i^p$ are zero, we can remove the state $X_i^p$ from the state space of the node $X_i$; we will see such an example in Section 5).

If the rank of a matrix is $r$, it can be represented as a product $\mathbf{Q}^T \mathbf{R}$ of two matrices[2]: matrix $\mathbf{Q}$ of size $r \times |X_i|$ and matrix $\mathbf{R}$ of size $r \times |X_j|$. So, we can represent $|X_1| \times |X_2|$ numbers for a sensitivity $\mathbf{S}_{ij}$ by $(|X_1| + |X_2|) \times \text{rank}(\mathbf{S}_{ij})$ numbers for the matrices $\mathbf{Q}_{ij}$ and $\mathbf{R}_{ij}$ (from them $(\text{rank}(\mathbf{S}_{ij})+1) \times \text{rank}(\mathbf{S}_{ij})$ numbers are still redundant due to the linear dependencies between the rows of the matrices). We prove properties of the sensitivities in Section 3.2 which will allow us to do probabilistic inference in tree networks in Section 4.

Many exact algorithms are based on the representation of a network as a tree. Any belief network can be converted to a tree of multiply-valued nodes by, for example, a breadth-first search of the moralized graph. A simple representation of a conditional probability matrix between two nodes $X_i$ and $X_j$ in the tree would have size $|X_i| \times |X_j|$. Such a matrix is likely to be much larger than necessary, owing to the loss of some information about conditional independence in the tree representation compared to the original representation in (2). For example, given two multiply-valued nodes $X_i$ and $X_j$ in the tree of multiply-valued nodes, the conditional probability $p(X_i|X_j)$ often does not depend on some of the nodes in the compound node $X_j$. In this case, every other column in the conditional probability matrix is the same and the rank is low. This means that the QR-representation is space-efficient compared to the full-matrix representation and recovers conditional independence information lost in the process of a network conversion to a tree form. In practice, since the size of an average multiply-valued node is about ten simple nodes and a large enough part of them is conditionally independent of another multiply-valued node given some other simple nodes, the rank is as a rule order of magnitude lower than the dimensions of the matrix.

### 3.2 PROPERTIES OF SENSITIVITIES

**Lemma 1** *If a node $X_i$ is conditionally independent of a node $X_k$ given a node $X_j$, $\mathbf{CI}(X_i, X_j, X_k)$, the sensitivity $\mathbf{S}_{ik}$ of node $X_i$ with respect to node $X_k$ is a product of the sensitivities $\mathbf{S}_{ij}$ of the node $X_i$ with respect to the node $X_j$, and $\mathbf{S}_{jk}$ of the node $X_j$ with*

---

[2] Although it is not essential in this paper, the matrix $\mathbf{Q}$ can be made an orthonormal matrix and the matrix $\mathbf{R}$ a triangular matrix with all entries below the leading diagonal zero.

respect to the node $X_k$:

$$\mathbf{S}_{ik} = \mathbf{S}_{ij}\mathbf{S}_{jk}. \tag{6}$$

*The rank of the $\mathbf{S}_{ik}$ matrix is less than or equal to the rank of either the $\mathbf{S}_{ij}$ or $\mathbf{S}_{jk}$ matrices.*

**Proof:** From the definition of conditional independence $p(X_i|X_j, X_k) = p(X_i|X_j)$:

$$\begin{aligned}p(X_i^p|X_k^q) &\equiv \sum_t p(X_i^p|X_j^t, X_k^q)p(X_j^t|X_k^q)\\ &= \sum_t p(X_i^p|X_j^t)p(X_j^t|X_k^q).\end{aligned}$$

The product of the sensitivities $\mathbf{S}_{ij}$ and $\mathbf{S}_{jk}$ is:

$$\begin{aligned}(\mathbf{S}_{ij}\mathbf{S}_{jk})_{pq} &= \sum_t \left(p(X_i^p|X_j^t) - \frac{\sum_s p(X_i^p|X_j^s)}{|X_j|}\right)(\mathbf{S}_{jk})_{tq}\\ &= \sum_t p(X_i^p|X_j^t)(\mathbf{S}_{jk})_{tq} - \\ &\quad \frac{\sum_s p(X_i^p|X_j^s)}{|X_j|}\sum_t (\mathbf{S}_{jk})_{tq}\\ &= \sum_t p(X_i^p|X_j^t)p(X_j^t|X_k^q) - \\ &\quad \sum_t p(X_i^p|X_j^t)\frac{\sum_s p(X_j^t|X_k^s)}{|X_k|}\\ &= p(X_i^p|X_k^q) - \frac{\sum_{t,s} p(X_i^p|X_j^t)p(X_j^t|X_k^s)}{|X_k|}\\ &= p(X_i^p|X_k^q) - \frac{\sum_s p(X_i^p|X_k^s)}{|X_k|} = (\mathbf{S}_{ik})_{pq}.\end{aligned}$$

The sum $\sum_t (\mathbf{S}_{jk})_{tq}$ is zero since the vector $(1, 1, \ldots, 1)$ is in the left nullspace of the sensitivity matrix $\mathbf{S}_{jk}$. The last statement of the Lemma about ranks follows from the linear algebra statement that the rank of a product of two matrices is less than or equal to the rank of each one of them. □

To estimate the computational complexity of the transformation (6), we represent the sensitivities as a product of $\mathbf{Q}$ and $\mathbf{R}$ matrices, i.e. $\mathbf{S}_{ij} = \mathbf{Q}_{ij}^T\mathbf{R}_{ij}$. The product $\mathbf{Z}_{ijk} = \mathbf{R}_{ij}\mathbf{Q}_{jk}^T$ and the product $\mathbf{Z}_{ijk}\mathbf{R}_{jk}$ take $(|X_j| \times \operatorname{rank}(\mathbf{S}_{ij}) \times \operatorname{rank}(\mathbf{S}_{jk}))$ and $(|X_k| \times \operatorname{rank}(\mathbf{S}_{ij}) \times \operatorname{rank}(\mathbf{S}_{jk}))$ multiplications and about the same number of summations correspondingly. The sensitivity $\mathbf{S}_{ik}$ is then given by a product $\mathbf{Q}_{ij}^T(\mathbf{Z}_{ijk}\mathbf{R}_{jk})$, which is a QR-representation with matrices $\mathbf{Q}_{ik} = \mathbf{Q}_{ij}$ and $\mathbf{R}_{ik} = \mathbf{Z}_{ijk}\mathbf{R}_{jk}$. Thus, the computational complexity of the node reduction (6) is $O(((|X_j| + |X_k|) \times \operatorname{rank}(\mathbf{S}_{ij}) \times \operatorname{rank}(\mathbf{S}_{jk}))$ in the QR-representation, as opposed to $O(|X_i||X_j||X_k|)$ in the full-matrix representation.

**Lemma 2** *The ranks of the sensitivity $\mathbf{S}_{ij}$ of a node $X_i$ with respect to node $X_j$ and the sensitivity $\mathbf{S}_{ji}$ of a node $X_j$ with respect to node $X_i$ are equal. The transformation from the $\mathbf{S}_{ij}$ to $\mathbf{S}_{ji}$ in the QR-representation has the form:*

$$\begin{aligned}\mathbf{Q}_{ji} &= \mathbf{R}_{ij}(\mathbf{\Lambda}_j - \mathbf{P}_j\mathbf{P}_j^T);\\ \mathbf{R}_{ji} &= \mathbf{Q}_{ij}\mathbf{\Lambda}_i^{-1}(\mathbf{I} - \frac{1}{|X_i|}\mathbf{E}).\end{aligned} \tag{7}$$

**Proof:** Since

$$\begin{aligned}p(X_i^p) &\equiv \sum_q p(X_i^p|X_j^q)p(X_j^q)\\ &= \sum_q \left((\mathbf{S}_{ij})_{pq} + \frac{\sum_s p(X_i^p|X_j^s)}{|X_j|}\right)p(X_j^q)\\ &= \sum_q (\mathbf{S}_{ij})_{pq}p(X_j^q) + \frac{\sum_s p(X_i^p|X_j^s)}{|X_j|},\end{aligned}$$

the conditional probability $p(X_i|X_j)$ as well as the conditional probability $p(X_j|X_i)$ can be expressed in terms of the sensitivity $\mathbf{S}_{ij}$ of the node $X_i$ with respect to the node $X_j$ and the probability distributions of nodes $X_i$ and $X_j$:

$$p(X_i^p|X_j^q) = (\mathbf{S}_{ij})_{pq} + p(X_i^p) - \sum_q (\mathbf{S}_{ij})_{pq}p(X_j^q) \tag{8}$$

and

$$\begin{aligned}p(X_j^q|X_i^p) &\equiv p(X_i^p|X_j^q)p(X_j^q)/p(X_i^p)\\ &= \left((\mathbf{S}_{ij})_{pq} + p(X_i^p) - \sum_q (\mathbf{S}_{ij})_{pq}p(X_j^q)\right)\frac{p(X_j^q)}{p(X_i^p)}.\end{aligned} \tag{9}$$

From the conditional probability $p(X_j|X_i)$ we can find the sensitivity $\mathbf{S}_{ji}$ by eliminating corresponding subspaces using (5). The rank of the conditional probability matrix is always one greater than the rank of the sensitivity matrix; multiplication of any row/column by a positive constant $p(X_j^q)$ or $1/p(X_i^p)$ does not change the rank of a matrix; converting the conditional probability to the sensitivity always decreases the rank by one.

In matrix notations, the transformation from $\mathbf{S}_{ij}$ to $\mathbf{S}_{ji}$ is:

$$\mathbf{S}_{ji} = \mathbf{\Lambda}_j\left(\mathbf{S}_{ij} + (\mathbf{P}_i - \mathbf{S}_{ij}\mathbf{P}_j)\mathbf{E}\right)^T \mathbf{\Lambda}_i^{-1}(\mathbf{I} - \frac{1}{|X_i|}\mathbf{E}).$$

The product $\mathbf{P}_i^T\mathbf{\Lambda}_i^{-1}$ is $\mathbf{E}$; $\mathbf{E}(\mathbf{I} - \frac{1}{|X_i|}\mathbf{E})$ is zero; and $\mathbf{\Lambda}_j\mathbf{E}\mathbf{P}_j^T$ is $\mathbf{P}_j\mathbf{P}_j^T$. After elimination of the zero term the above expression is:

$$\mathbf{S}_{ji} = (\mathbf{\Lambda}_j - \mathbf{P}_j\mathbf{P}_j^T)\mathbf{S}_{ij}^T\mathbf{\Lambda}_i^{-1}(\mathbf{I} - \frac{1}{|X_i|}\mathbf{E}).$$

The transformation in the QR-representation has the form:

$$\mathbf{Q}_{ji}^T\mathbf{R}_{ji} = (\mathbf{\Lambda}_j - \mathbf{P}_j\mathbf{P}_j^T)\mathbf{R}_{ij}^T\mathbf{Q}_{ij}\mathbf{\Lambda}_i^{-1}(\mathbf{I} - \frac{1}{|X_i|}\mathbf{E}),$$

which is equivalent to (7). □

An important corollary of Lemma 2 is that the sensitivity matrix $\mathbf{S}_{ij}$ and probability distributions $\mathbf{P}_i$ and $\mathbf{P}_j$ for nodes $X_i$ and $X_j$ uniquely define conditional probabilities $p(X_i|X_j)$ and $p(X_j|X_i)$.

Taking into account a special form of the matrices in (7), the transformation from $\mathbf{Q}_{ij}$ to $\mathbf{R}_{ji}$ can be computed in $O(|X_i| \times \operatorname{rank}(\mathbf{S}_{ij}))$, and the transformation from $\mathbf{R}_{ij}$ to $\mathbf{Q}_{ji}$ in $O(|X_j| \times \operatorname{rank}(\mathbf{S}_{ij}))$ time. As a result, the computational complexity of the sensitivity reversal (7) is $O(((|X_i| + |X_j|) \times \operatorname{rank}(\mathbf{S}_{ij}))$ in the QR-representation, as opposed to $O(|X_i||X_j|)$ in the full-matrix representation.

**Lemma 3** *If a node $X_i$ is conditionally independent of a node $X_k$ given a node $X_j$, $\mathrm{CI}(X_i, X_j, X_k)$, then, for any instantiation of the node $X_k$, the posterior sensitivity $\mathbf{S}_{ij}^{(1)}$ of the node $X_i$ with respect to the node $X_j$ after instantiation is the same as the prior sensitivity $\mathbf{S}_{ij}^{(0)}$, and the change $\Delta p(X_i) \equiv p^{(1)}(X_i) - p^{(0)}(X_i)$ in the probability $p(X_i)$ of the node $X_i$ is the sensitivity $\mathbf{S}_{ij}$ of node $X_i$ with respect to node $X_j$ multiplied by the change $\Delta p(X_j) \equiv p^{(1)}(X_j) - p^{(0)}(X_j)$ in the probability $p(X_j)$ of node $X_j$:*

$$\Delta p(X_i^p) = \sum_q (\mathbf{S}_{ij})_{pq} \Delta p(X_j^q). \tag{10}$$

**Proof:** The proof of this lemma follows directly from conditional independence: $p(X_i|X_j, X_k) = p(X_i|X_j)$. Any instantiation of the node $X_k$ does not change the values for the conditional probability $p(X_i|X_j)$. The change in the probability $p(X_i)$ of the node $X_i$ is:

$$\begin{aligned}
\Delta p(X_i^p) &\equiv p^{(1)}(X_i^p) - p^{(0)}(X_i^p) \\
&= \sum_q p(X_i^p|X_j^q) p^{(1)}(X_j^q) - \\
&\quad \sum_q p(X_i^p|X_j^q) p^{(0)}(X_j^q) \\
&= \sum_q p(X_i^p|X_j^q) \left( p^{(1)}(X_j^q) - p^{(0)}(X_j^q) \right) \\
&= \sum_q p(X_i^p|X_j^q) \Delta p(X_j^q),
\end{aligned}$$

which is equivalent to (10) considering that the change $\Delta p(X_j^q)$ in the probability distribution of the node $X_j$ is orthogonal to the vector $(1, 1, \ldots, 1)$. $\square$

The last lemma gives a rule to update probabilities given a single instantiated node. We have to propagate the change throughout the nodes ordered according to the property of conditional independence. In Section 4 we show that a tree network has a set of very convenient properties which allow to do probabilistic inference by updating sensitivities and probability distributions without ever converting the sensitivities to conditional probabilities. Let us look at the very important case of binary sensitivities first.

### 3.3 BINARY SENSITIVITIES

Since every multiply-valued node can be represented as a collection of binary nodes, binary nodes are very important for the theory of belief networks. The *binary sensitivity* of a binary node $x_i$ with respect to a binary node $x_j$ is a scalar $S_{ij}$:

$$\begin{aligned}
S_{ij} &\equiv p(i|j) - p(i|\bar{j}) \\
&= p(\bar{i}|\bar{j}) - p(\bar{i}|j).
\end{aligned} \tag{11}$$

As a matrix, the sensitivity $\mathbf{S}_{ij}$ for binary nodes is:

$$\begin{aligned}
\mathbf{S}_{ij} &= (-1/\sqrt{2}\ \ 1/\sqrt{2})^T S_{ij} (-1/\sqrt{2}\ \ 1/\sqrt{2}) \\
&= S_{ij} \left(\mathbf{I} - \tfrac{1}{2}\mathbf{E}\right).
\end{aligned} \tag{12}$$

Let us consider the two basic operations of probabilistic inference in light of this approach which uses sensitivities rather than conditional probabilities. Node reduction (3) corresponds to multiplying two sensitivities $S_{32}$ and $S_{21}$ to find sensitivity $S_{31}$ or sensitivities $S_{12}$ and $S_{23}$ to find sensitivity $S_{13}$, for example:

$$\begin{aligned}
\mathbf{S}_{31} &= S_{31}(\mathbf{I} - \tfrac{1}{2}\mathbf{E}) = \mathbf{S}_{32}\mathbf{S}_{21} \\
&= S_{32}(\mathbf{I} - \tfrac{1}{2}\mathbf{E}) S_{21}(\mathbf{I} - \tfrac{1}{2}\mathbf{E}) \\
&= S_{32} S_{21}(\mathbf{I} - \tfrac{1}{2}\mathbf{E})^2 \\
&= S_{32} S_{21}(\mathbf{I} - \tfrac{1}{2}\mathbf{E}),
\end{aligned} \tag{13}$$

and the number of operations to reduce a node is one multiplication, as compared with 8 multiplications and 4 summations using conditional probabilities directly (see (3)).

Arc reversal (4) corresponds to reversing sensitivity $\mathbf{S}_{13}$ to $\mathbf{S}_{31}$; after substitution of (12) into (7), we have for binary sensitivities:

$$S_{ji} = \frac{p(\bar{j})p(j)}{p(\bar{i})p(i)} S_{ij}, \tag{14}$$

and the number of operations to reverse an arc is 3 multiplications and one division, as compared with 4 multiplications, 2 summations, and 4 divisions using conditional probabilities directly (see (4)).

If we have the binary sensitivity $S_{ij}$ of a node $x_i$ with respect to a node $x_j$, probabilistic inference $p(i|x_j)$ is reduced to one summation and one multiplication:

$$p^{(1)}(i) = p^{(0)}(i) + S_{ij} \Delta p(j), \tag{15}$$

where $\Delta p(j) = p^{(1)}(j) - p^{(0)}(j)$ and is either $(-p^{(0)}(j))$, for the instantiation to *false*, or $p^{(0)}(\bar{j})$, for the instantiation to *true*.

The binary sensitivity is always less than one. Since in a chain of nodes sensitivities are multiplied by each other according to (13), we can expect that in a binary tree network the sensitivity between nodes decreases exponentially with the distance between the nodes. This suggests that to evaluate a query, we may be able to neglect instantiated nodes that are far enough from the query node. In fact, it is possible to prove the following theorem [Kozlov and Singh, 1995]:

**Theorem 1** *Given a belief network represented by a tree of binary nodes; a set of prior binary sensitivities between nodes $S_{ij}^{(0)}$ : $\left|S_{ij}^{(0)}\right| < \alpha < 1$; and a set of prior probability distributions of the nodes $\{p^{(0)}(\bar{i}), p^{(0)}(i)\}$ : $0 < \eta < p^{(0)}(\bar{i}) p^{(0)}(i)$, approximate probabilistic inference for a single query node and a set of instantiated nodes $I$ of size $N(I)$ is possible in time independent of the size of the network. The relative error of the result is guaranteed to be less than $(\exp(\epsilon) - 1)$ for any $\epsilon$ : $0 < \epsilon < 1$; the computational complexity of the algorithm is $O\left((N(I))^2 \log_\alpha(\eta\epsilon/2N(I))\right)$.*

The theorem can be easily extended to the trees with a finite number of multiply-valued nodes and a bound on the number of states per node, as well as polytrees with a finite number of nodes with multiple parents and a bound on the number of parents, since these nodes can be reduced in a finite amount of time. The theorem shows that for a certain class of networks we can do approximate probabilistic inference with the *exact* upper bound on the error of the result in time that *does not depend* on the size of the network.

## 4 PROBABILISTIC INFERENCE

According to Lemma 3, the change in the probability of a node can be propagated through the sets of conditionally independent nodes. In this section we take advantage of the topological properties of a tree network and give two algorithms for an exact probabilistic inference in a tree network based on updating sensitivities between nodes and probability distributions of the nodes.

The first algorithm, in Section 4.1, evaluates the new sensitivities between nodes and the new probability distributions of nodes after a single node has been instantiated. It uses a topological property of a tree network that each node in the tree network can be made the root of the tree by reversing all edges from the node up to the original root. We make the instantiated node the root of the tree. Since all descendants of a node are conditionally independent of the root given the node itself, we can recursively apply (10) to evaluate the change in the probability distributions throughout the network. The second algorithm, in Section 4.2, handles multiple instantiated nodes but a single query node at a time. It uses the topological property of a tree network that the instantiation of any node $X$ in the tree divides the whole network into $N(\mathbf{Nb}(X))$ conditionally independent parts, where $\mathbf{Nb}(X)$ is the set of the node $X$ neighbors.

We describe our algorithms in the object-oriented paradigm where a potential user or a node sends messages to other nodes with requests for actions. Each node $X_j$ stores a part of the sensitivity matrix $\mathbf{S}_{ij}$ of all its neighbors $X_i$ with respect to the node itself, the $\mathbf{R}_{ij}$ matrix.[3] The updates are sent from a node to an adjacent node and represent a product $\mathbf{Y} = \mathbf{R}_{ij}\Delta\mathbf{P}_j$ (we need to send only rank($\mathbf{S}_{ij}$) numbers; we claim that this is the absolute minimum of information a node $X_j$ has to send to a node $X_i$ during a process of probabilistic inference). The left part of the product $\Delta\mathbf{P}_i = \mathbf{Q}_{ij}^T\mathbf{Y}$ is computed by the node $X_i$. The computational complexity of one update is this scheme is $O\bigl((|X_i| + |X_j|) \times \text{rank}(\mathbf{S}_{ij})\bigr)$.

---

[3]Since there are infinitely many matrices $\mathbf{Q}_{ij}$ and $\mathbf{R}_{ij}$ that give $\mathbf{S}_{ij} = \mathbf{Q}_{ij}^T\mathbf{R}_{ij}$, the matrices $\mathbf{R}_{ij}$ and $\mathbf{R}_{ji}$ should be mutually consistent: one of them should be the transformation (7) of the corresponding $\mathbf{Q}$ matrix.

### 4.1 A SINGLE INSTANTIATED NODE AND MULTIPLE QUERY NODES

```
method instantiate() {
   update P_this to the instantiated value;
   for each X_below from Nb(X_this) {
      X_below → simq(X_this, R_below this (P_this − P_this^(0)));
   }
}

method simq(X_above, Y) {
   Q_this above = R_above this (Λ_this − P_this P_this^T);
   P_this = P_this^(0) + Q_this above^T Y;
   R_above this = Q_this above Λ_this^{-1} (I − (1/|X_this|)E);
   for each X_below from (Nb(X_this) − X_above) {
      X_below → simq(X_this, R_below this (P_this − P_this^(0)));
   }
}
```

Figure 1: Algorithm for a Single Instantiated Node and Multiple Query Nodes

The inference algorithm for a single instantiated node and multiple query nodes (**simq**) is shown in Figure 1. It begins with sending a message **instantiate()** to the instantiated node. The node instantiates itself[4] and sends a product $\mathbf{Y} = \mathbf{R}\Delta\mathbf{P}$ to each of its neighbors through the method **simq()**. At each neighbor, this method computes the $\mathbf{Q}$ matrix using (7), transforms the message $\mathbf{Y}$ from the other node to the change in the probability $\Delta\mathbf{P} = \mathbf{Q}^T\mathbf{Y}$, and computes the updated $\mathbf{R}$ matrix using (7) with the updated probability distribution. Thus, before the *for each* loop, the node has the updated posterior probability and the updated posterior $\mathbf{R}$ matrix. The *for each* loop recursively sends updates to the neighbors in the depth-first search order through the method **simq()**. After the return of the original call to **instantiate()**, the process can be repeated for another node instantiation after setting $\mathbf{P}^{(0)}$ to $\mathbf{P}$ in each of the nodes.

### 4.2 MULTIPLE INSTANTIATED NODES AND A SINGLE QUERY NODE

With multiple instantiated nodes, given that the set $I$ of the instantiated nodes is finite, we could apply incremental instantiation and repeat the algorithm in Figure 1 to evaluate the probability distributions of the nodes in the network. However, for a single query node, we can perform probabilistic inference in time independent of the number of instantiated nodes $N(I)$. The second algorithm is shown in Figure 2. Note that unlike the LS algorithm, this algorithm guarantees to update the probability distribution of the query node only, and not all the nodes in the network.

---

[4]The update of the probability distribution of a node can be extended to the case of the instantiation of simple nodes in a compound node.

```
method query(I) {
    X_this → markBarren(X_this, I);
    for each not barren X_below from Nb(X_this) {
        Y_below = X_below → misq(X_this, I,
                        R_below this (P_this − P^(0)_this));
        Q_this below = R_below this (Λ_this − P_this P^T_this);
        P_this = P_this + Q^T_this below Y_below;
    }
    if (X_this ∈ I) update P_this to the instantiated value;
}

method misq(X_above, I, Y_above) {
    if (X_this ∈ I or (Nb(X_this) − X_above) has more
                    than one not barren node) {
        Q_this above = R_above this (Λ_this − P_this P^T_this);
        P_this = P^(0)_this + Q^T_this above Y_above;
        R_above this = Q_this above Λ^{-1}_this (I − 1/|X_this| E);
        P^(1)_this = P_this;
        for each not barren X_below from
                    (Nb(X_this) − X_above) {
            Y_below = X_below → misq(X_this, I,
                            R_below this (P_this − P^(0)_this));
            Q_this below = R_below this (Λ_this − P_this P^T_this);
            P_this = P_this + Q^T_this below Y_below;
        }
        if (X_this ∈ I) update P_this to the instantiated value;
        return R_above this (P_this − P^(1)_this);
    } else {
        set X_below to
            a not barren node from (Nb(X_this) − X_above);
        Q_this above = R_above this (Λ_this − P_this P^T_this);
        Z_below this above = R_below this Q^T_this above;
        Y_below = X_below → misq(X_this, I,
                        Z_below this above Y_above);
        return Z^T_below this above Y_below;
    }
}
```

Figure 2: Algorithm for Multiple Instantiated Nodes and a Single Query Node

The probabilistic inference begins with sending a message $\text{query}(I)$ to the query node. Method $\text{query}()$ first marks all barren nodes with the recursive method $\text{markBarren}()$ (not shown). A barren node is a node which is not instantiated and does not have instantiated descendants; these nodes can be safely removed from the graph [Shachter, 1986] by a recursive algorithm. Then, message $\text{misq}()$ is sent recursively to all of the neighbors. Upon reception of the message $\text{misq}()$, a node either evaluates the updated probability or, if it is not instantiated and has only two not barren neighbors, reduces itself.

Let us see why this algorithm works. The method $\text{misq}()$ has three sets of probability distributions: the prior probability distribution, the current probability distribution and the probability distribution before the $\textit{for each}$ loop. A node $X_{\text{this}}$ divides the whole network into $N(\text{Nb}(X_{\text{this}}))$ conditionally independent parts. On the entry to the method $\text{misq}()$, instantiations could have been made only in one of them, the part to which the node $X_{\text{this}}$ is connected through the node $X_{\text{above}}$. By Lemma 3 sensitivities of all nodes from the set $(\text{Nb}(X_{\text{this}}) - X_{\text{above}})$ with respect to the node $X_{\text{this}}$ have their prior value. Instantiation of the nodes in one of these parts still preserves sensitivities of the other parts with respect to $X_{\text{this}}$ at their prior value. The changes in the probability distribution are accumulated and distributed to the other parts. When the $\textit{for each}$ loop exits, the node $X_{\text{this}}$ instantiates itself and returns the information about the change in the probability to the node $X_{\text{above}}$. The node $X_{\text{above}}$, in its turn, accumulates the changes from all of its neighbors. Since each edge can be traversed only twice in the depth-first search, the computation time is bounded by the size of the network. Let us illustrate the process of probabilistic inference based on sensitivities with examples.

## 5 EXAMPLES

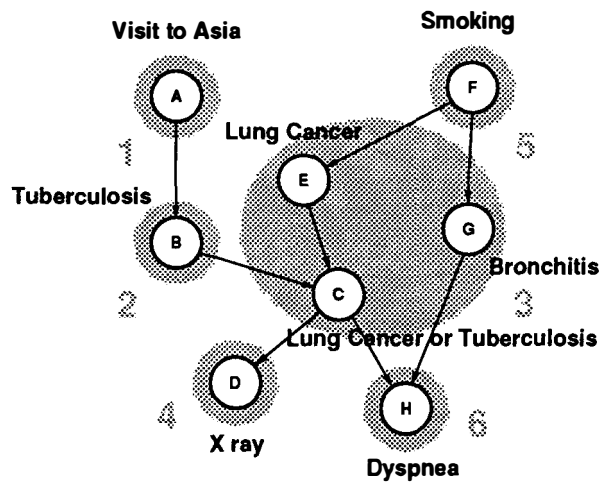

Figure 3: The Asia Network

We demonstrate the usage of sensitivities on the well known Asia network [Lauritzen and Spiegelhalter, 1988] shown in Figure 3. The nodes in the network are: $x_A$ – "Visit to Asia", $x_B$ – "Tuberculosis, $x_C$ – "Lung Cancer or Tuberculosis", $x_D$ – "X-ray", $x_E$ – "Lung Cancer", $x_F$ – "Smoking", $x_G$ – "Bronchitis", $x_H$ – "Dyspnea". Tuberculosis and lung cancer are equally likely to cause shortness of breath (dyspnea), and are also equally likely to cause a positive chest X-ray. Bronchitis is another cause of dyspnea. Tuberculosis is more likely in people who have visited Asia. Smoking can cause lung cancer and bronchitis. The belief network describing this probabilistic model is given in Figure 3.

To convert the network to a multiply-valued node tree representation we combined nodes $x_C$, $x_E$, and $x_G$ in

$$\mathbf{S}_{21} \quad \begin{pmatrix} -1/\sqrt{2} & 1/\sqrt{2} \end{pmatrix}^T \begin{pmatrix} -.0400/\sqrt{2} & .0400/\sqrt{2} \end{pmatrix}$$

$$\mathbf{S}_{32} \quad \begin{pmatrix} -.5535 & -.4400 & .5535 & .4400 & 0 & 0 \end{pmatrix}^T \begin{pmatrix} -.6726/\sqrt{2} & .6726/\sqrt{2} \end{pmatrix}$$

$$\mathbf{S}_{43} \quad \begin{pmatrix} -1/\sqrt{2} & 1/\sqrt{2} \end{pmatrix}^T \begin{pmatrix} -.8768 & -.8768 & .4384 & .4384 & .4384 & .4384 \end{pmatrix}$$

$$\mathbf{S}_{53} \quad \begin{pmatrix} -1/\sqrt{2} & 1/\sqrt{2} \end{pmatrix}^T \begin{pmatrix} -.4069 & .0220 & -.4069 & .0220 & .3132 & .4565 \end{pmatrix}$$

$$\mathbf{S}_{63} \quad \begin{pmatrix} -1/\sqrt{2} & 1/\sqrt{2} \end{pmatrix}^T \begin{pmatrix} -.8250 & .1650 & .0236 & .3064 & .0236 & .3064 \end{pmatrix}$$

one multiply-valued node $X_3$. The six multiply-valued nodes are denoted by shadows on the background (all nodes but $X_3$ remain binary). An instantiation of the node $X_2$ divides the whole network into two, and an instantiation of the node $X_3$ into four conditionally independent parts. The conditional probabilities for this network can be found in [Neapolitan, 1990]. The probability distributions and conditional probabilities for the multiply-valued nodes were computed using the LS algorithm and a perl script. The conditional probabilities were converted to sensitivities using MATLAB.[5] Starting from this preprocessed information, the rest of the operations in the examples can be performed with a hand calculator.

Table 1: Prior Probability Distributions of Nodes in the Asia Network

| Node | | $X^0$ | $X^1$ | $X^2$ | $X^3$ | $X^4$ | $X^5$ | $X^6$ | $X^7$ |
|---|---|---|---|---|---|---|---|---|---|
| $X_1$ | $\{x_A\}$ | .9900 | .0100 | | | | | | |
| $X_2$ | $\{x_B\}$ | .9896 | .0104 | | | | | | |
| $X_3$ | $\{x_C, x_E, x_G\}$ | .5210 | .4141 | 0 | 0 | .0055 | .0044 | .0235 | .0315 |
| $X_4$ | $\{x_D\}$ | .8897 | .1103 | | | | | | |
| $X_5$ | $\{x_F\}$ | .5000 | .5000 | | | | | | |
| $X_6$ | $\{x_H\}$ | .5640 | .4360 | | | | | | |

Table 2: Sensitivities between Adjacent Nodes in the Asia Network in QR-representation

The probability distributions for nodes $X_1$ through $X_6$ are given in Table 1. The states (columns) are enumerated from zero, zero being the state where all nodes are *false*. Different states of the compound node are enumerated according to expression $\sum_i 2^{i-1} x_i$, where $i$ is the position of a node from the right. Thus, for the compound node $X_3 = \{x_C, x_E, x_G\}$ the ordering of states is $\overline{C}\,\overline{E}\,\overline{G}, \overline{C}\,\overline{E}\,G, \ldots, C\,E\,G$. From the table we can see that the probability of states $X_3^2$ and $X_3^3$ is zero. Checking the conditional probabilities we find that indeed $p(\overline{C}|E) = 0$ is zero and these states are impossible in a universe described by this belief network. Since their probability is *always* zero, we discard these states from further consideration. The sensitivities of the adjacent nodes in the QR-representation are given in Table 2. The information in Tables 1 and 2 completely describes the Asia network.

To illustrate our method we solve two practical problems. We perform probabilistic inference using node reduction ((6) and (13)), sensitivity reversal ((7) and (14)), and update rule ((10) and (15)). We apply incremental instantiation and update sensitivities and probability distributions between any two consecutive instantiations if necessary.

---

[5]The transformations used in Sections 4 and 5 were implemented in MATLAB and are available upon request.

**Example 1** *From Table 1 we can see that among the general population about 1% visited Asia. What is the increase in this percentage for people with dyspnea?*

The strategy to answer this query is to compute the binary sensitivity (Section 3.3) $S_{HA}$ and to convert it using (14) to the binary sensitivity $S_{AH}$. The sensitivity $\mathbf{S}_{61}$ of node $X_6$ with respect to node $X_1$ is:

$$\mathbf{S}_{61} = \mathbf{S}_{63}\mathbf{S}_{32}\mathbf{S}_{21} = \mathbf{Q}_{63}^T \mathbf{R}_{63} \mathbf{Q}_{32}^T \mathbf{R}_{32} \mathbf{Q}_{21}^T \mathbf{R}_{21}.$$

Since matrices $\mathbf{R}_{63}$ and $\mathbf{Q}_{32}$ contain only one row, the product $\mathbf{R}_{63}\mathbf{Q}_{32}^T$ is an inner product of two vectors of size six and, taking into account two zeros in the $\mathbf{R}_{63}$ matrix, it takes 4 multiplications and 3 summations to evaluate: $\mathbf{R}_{63}\mathbf{Q}_{32}^T = .5319$. The product $\mathbf{R}_{32}\mathbf{Q}_{21}^T$ is an inner product of two vectors of size two and, taking into account the symmetric form of $\mathbf{R}_{32}$ and $\mathbf{Q}_{21}$, we evaluate: $\mathbf{R}_{32}\mathbf{Q}_{21}^T = .6726$. For the binary sensitivity in the form (12), we get:

$$\mathbf{S}_{61} = .5319 \times .6726 \times .0400 \times \left(\mathbf{I} - \tfrac{1}{2}\mathbf{E}\right)$$
$$= .01431 \times \left(\mathbf{I} - \tfrac{1}{2}\mathbf{E}\right).$$

Now, we reverse the binary sensitivity $S_{HA}$:

$$S_{AH} = \frac{p(\overline{A})p(A)}{p(\overline{H})p(H)} S_{HA} = \frac{.009900}{.2459} .01431 \approx 5.8 \times 10^{-4}.$$

Rephrasing the result, the probability $p(A)$ of the node $x_A$ changes by $\approx 5.8 \times 10^{-4}$ each time when the probability $p(H)$ of the node $x_H$ changes by one. If node $x_H$ is instantiated to *true*, the change in the probability $p(H)$ is $\Delta p(H) = p(\overline{H}) = .5640$ and the change in the probability $p(A)$ of the node $x_A$ is:

$$\Delta p(A) = .5640 \times 5.8 \times 10^{-4} = 3.3 \times 10^{-4}.$$

**Answer:** The increase in the percentage is $\Delta p(x_A) = 3.3 \times 10^{-2}\%$.

**Example 2** *A person visited Asia and has a positive chest X-ray result. What is the probability of dyspnea?*

We shall perform incremental instantiation with the set $\{x_A, x_D\}$. We choose to instantiate node $x_A$ first. Using (6), we calculate $\mathbf{S}_{31}$, $\mathbf{S}_{41}$, and $\mathbf{S}_{61}$ (see the previous example):

$$\mathbf{S}_{31} = \mathbf{Q}_{32} \times .6726 \times \mathbf{R}_{21};$$
$$\mathbf{S}_{41} = .03515 \times (\mathbf{I} - \tfrac{1}{2}\mathbf{E});$$
$$\mathbf{S}_{61} = .01431 \times (\mathbf{I} - \tfrac{1}{2}\mathbf{E}).$$

The change in the probability distribution of the node $X_1$ after instantiation is $\Delta \mathbf{P}_1 = (-.9900 \ .9900)^T$. Multiplying the sensitivity matrices by the change in the probability distribution of the node $X_1$ gives the changes in the probability distribution of nodes $X_3$, $X_4$, and $X_6$. Adding the changes to the probability distributions in Table 1, we get:

$$\mathbf{P}_3^{(1)} = (.5002 \ .3975 \ .0263 \ .0210 \ .0235 \ .0315)^T;$$
$$\mathbf{P}_4^{(1)} = (.8549 \ .1451)^T;$$
$$\mathbf{P}_6^{(1)} = (.5498 \ .4502)^T.$$

Now we have to reverse the sensitivity $\mathbf{S}_{43}$ and to instantiate node $X_4$. The product $\mathbf{Q}_{43}(\mathbf{\Lambda}_4^{(1)})^{-1}$ is the $\mathbf{Q}_{43}$ matrix with every element divided by the corresponding probability from the probability distribution $\mathbf{P}_4^{(1)}$:

$$\mathbf{Q}_{43}(\mathbf{\Lambda}_4^{(1)})^{-1} = \frac{1}{\sqrt{2}}(-1.170 \ 6.892).$$

The multiplication by $(\mathbf{I} - \tfrac{1}{2}\mathbf{E})$ on the right is equivalent to subtracting an arithmetic average $(-1.170 + 6.892)/2 = 2.861$ from both elements:

$$\mathbf{R}_{34}^{(1)} = \mathbf{Q}_{43}(\mathbf{\Lambda}_4^{(1)})^{-1}\left(\mathbf{I} - \frac{1}{2}\mathbf{E}\right) = \frac{1}{\sqrt{2}}(-4.031 \ 4.031).$$

Reversing $\mathbf{R}_{43}$ to $\mathbf{Q}_{34}^{(1)}$ requires a little more work, but not much: the product $\mathbf{R}_{43}\mathbf{\Lambda}_3^{(1)}$ is the $\mathbf{R}_{43}$ matrix with every element multiplied by the corresponding probability from the probability distribution $\mathbf{P}_3^{(1)}$, the product $\mathbf{R}_{43}\mathbf{P}_3^{(1)}$ is the sum of all elements of the previous result and equals $-.7423$, and, as a reader can check, the final result can be expressed as $(\mathbf{Q}_{34}^{(1)})_p = ((\mathbf{R}_{43})_p + .7423)((\mathbf{P}_3^{(1)})^T)_p$. The computation gives:

$$\mathbf{Q}_{34}^{(1)} = (-.0673 \ -.0535 \ .0311 \ .0248 \ .0277 \ .0372),$$

and $\mathbf{S}_{64}^{(1)} = \mathbf{Q}_{63}^T \mathbf{R}_{63}(\mathbf{Q}_{34}^{(1)})^T \mathbf{R}_{34}^{(1)}$ is:

$$\mathbf{S}_{64}^{(1)} = 0.06708 \times 4.031 \times (\mathbf{I} - \tfrac{1}{2}\mathbf{E})$$
$$\approx .27 \times (\mathbf{I} - \tfrac{1}{2}\mathbf{E}).$$

If the node $x_D$ is instantiated to *true*, the change in the probability $p^{(1)}(D)$ is $\Delta p^{(1)}(D) = p^{(1)}(\overline{D}) = .8549$ and the posterior probability $p^{(2)}(H)$ is:

$$p^{(2)}(H) = p^{(1)}(H) + S_{HD}^{(1)} \Delta p^{(1)}(D)$$
$$= .4502 + .27 \times .8549 = .68.$$

**Answer:** The probability of dyspnea is $p^{(2)}(H) = .68$.

## DISCUSSION

Sensitivities provide a simple and intuitive description of the dependencies in a belief network. The term sensitivity has been used before in a number of ways, but never in a way formal enough to show that probabilistic inference can be performed based on sensitivities alone. Two definitions that are close in spirit to ours are in [Laskey, 1993], where sensitivity is a derivative of a target probability with respect to network parameters, and in decision analysis [Howard, 1990], where the sensitivity describes the dependence of the expected value of the utility function on the input parameters to a probabilistic model. There have been other definitions of sensitivity as well [Pearl, 1988]. In our paper sensitivity is a linear transformation from the change in the probability of one node to the change in the probability of another node. The change can be due to the change in the network parameters, a change in the input parameters, or instantiations.

A belief network can be described in terms of sensitivities between nodes and probability distributions of the nodes. Since the transformation from this representation to the traditional representation in terms of conditional probabilities and back is a one-one relation, both representations have the same expressive power. A person thinking in terms of probabilities asks the question "What is the probability of an event if certain facts are true?" A person thinking in terms of sensitivities asks "How does the probability distribution change if some facts change?" Given the transformation rules in this paper, these two questions are identical. We have shown how to make probabilistic inference based on sensitivities only, without ever converting them back to conditional probabilities. Each node updates its own probability distribution based on the messages it receives from neighbors. In doing so it is not necessary to propagate messages throughout the whole network if we are concerned only with nodes in some localized part of the network.

Sensitivity is a very efficient and powerful representation of dependencies. For binary nodes the node reduction is more than eight times more efficient. For multiply-valued nodes the efficiency depends on the rank of the probability matrix. The computational complexity changes from $O(|X_i||X_j||X_k|)$ to $O((|X_j| + |X_k|) \times \text{rank}(\mathbf{S}_{ij}) \times \text{rank}(\mathbf{S}_{jk}))$ for the node reduction and from $O(|X_i||X_j|)$ to $O((|X_i| + |X_j|) \times \text{rank}(\mathbf{S}_{ij}))$ for the arc reversal. We believe that for the compound nodes obtained after

the conversion of an arbitrary network to a tree network, the sensitivity matrices should have a low rank (see discussion at the end of Section 3.1).

One might argue that the efficiency of the operations with sensitivities is a result of the work done in transforming conditional probabilities to sensitivities. In fact, this is exactly the result we want to achieve, i.e. to represent a belief network in a form that is convenient for probabilistic inference. This representation can be computed once and then reused any number of times to evaluate different queries. The step of conversion from conditional probabilities to sensitivities is not absolutely necessary either: the learning algorithms that construct belief networks can be easily changed to train sensitivities instead of conditional probabilities.

Finally, although even approximate probabilistic inference is *NP*-hard for general belief networks [Dagum and Luby, 1993], in Section 3.3 we showed that for a certain class of belief networks approximate probabilistic inference with the *exact* upper bound on the error of the result can be performed in time *independent* of the size of the network. In fact, sensitivities open a path to a whole class of approximate algorithms. Given a belief network **BN** in tree representation we can build a belief network $\mathbf{BN}(r)$ where $r$ is the maximum rank among all sensitivity matrices. Using linear algebra methods we can first remove those dependencies from the sensitivity matrices which don't affect the joint probability distribution much, thus reducing $r$. The computation time for inference will decrease linearly with $r$, while the precision will have a weaker dependence provided that $r$ is close to the original maximum rank.

On the whole, we believe that the concept of sensitivity is useful for probabilistic inference in belief networks. The reader is also referred to [Kozlov and Singh, 1995] for a more comprehensive treatment in which we build our theory beginning with a very simple case.

### Acknowledgements

We thank the reviewers for valuable comments. We also thank Steve Hanks and Bruce D'Ambrosio for answering our many questions, Ross Shachter, Greg Provan, and Malcolm Pradhan for discussions, Adam Galper for his implementation of the LS algorithm, John Hennessy for his support and guidance, and ARPA for financial support under contract no. N00039-91-C-0138.